\documentclass[11pt,a4paper]{article}
\usepackage[hyperref]{acl2020}
\usepackage{times}
\usepackage{latexsym}

\usepackage{url}

\usepackage{amsmath}

\usepackage{booktabs}       
\usepackage{multirow}
\usepackage{amssymb}

\usepackage{amsfonts}       
\usepackage{color}
\usepackage{enumitem}
\usepackage{graphicx}
\usepackage{graphics}
\usepackage{epstopdf} 
\usepackage{subfig}
\usepackage{caption}

\aclfinalcopy 

\title{Towards Non-task-specific Distillation of BERT via Sentence \\ Representation Approximation}

\author{
Bowen Wu$^1$, Huan Zhang$^{2}$\thanks{* Equal contribution during the internship at Tencent.}, Mengyuan Li$^{2}$\footnotemark[1], Zongsheng Wang$^1$, Qihang Feng$^1$,\\ \textbf{Junhong Huang}$^1$, \textbf{Baoxun Wang}$^1$ \\
$^1$Platform and Content Group, Tencent \\
$^2$Peking University, Beijing, China \\
{\tt \small{jasonbwwu,jasoawang,careyfeng,vincenthuang,asulewang@tencent.com}}\\
{\tt \small{zhanghuan123,limengyuan@pku.edu.cn}}\\
}

\date{}

\begin{document}
\maketitle
\begin{abstract}


Recently, BERT has become an essential ingredient of various NLP deep models due to its effectiveness and universal-usability. 
However, the online deployment of BERT is often blocked by its large-scale parameters and high computational cost. 
There are plenty of studies showing that the knowledge distillation is efficient in transferring the knowledge from BERT into the model with a smaller size of parameters. 
Nevertheless, current BERT distillation approaches mainly focus on task-specified distillation, 
such methodologies lead to the loss of the general semantic knowledge of BERT for universal-usability. 
In this paper, we propose a sentence representation approximating oriented distillation framework that can distill the pre-trained BERT into a simple LSTM based model without specifying tasks. 
Consistent with BERT, our distilled model is able to perform transfer learning via fine-tuning to adapt to any sentence-level downstream task. 
Besides, our model can further cooperate with task-specific distillation procedures. 
The experimental results on multiple NLP tasks from the GLUE benchmark show that our approach outperforms other task-specific distillation methods or even much larger models, i.e., ELMO, with efficiency well-improved.

\end{abstract}

\section{Introduction}
\label{sec:intro}

As one of the most important progress in the Natural Language Processing field recently, 
the Bidirectional Encoder Representation from Transformers (BERT)~\cite{devlin2019bert} has been proved to be effective in improving the performances of various NLP tasks, 
by providing a powerful pre-trained language model based on large-scale unlabeled corpora. 
Very recent studies have shown that the capability of BERT can be further enhanced by utilizing deeper architectures 
or performing the pre-training on larger corpora with appropriate guidance~\cite{radford2019language,yang2019xlnet,liu2019roberta}.


Despite its strength in building distributed semantic representations of sentences and supporting various NLP tasks,
BERT holds a huge amount of parameters that raises the difficulty of conducting online deployment due to its unsatisfying computational efficiency. 
To address this issue, 
various studies have been done to utilize knowledge distillation~\cite{hinton2015distilling} for compressing BERT and meanwhile keep its semantic modeling capability as much as possible ~\cite{chia2019transformer,tsai2019small}.
The distilling methodologies include simulating BERT with a much smaller model (e.g., LSTM)~\cite{tang2019distilling} 
and reducing some of the components, such as transformers, attentions, etc., to obtain the smaller model~\cite{sun2019patient,barkan2019scalable}. 

Nevertheless, the current methods highly rely on a labeled dataset upon a specified task.
Firstly, BERT is fine-tuned on the specified task to get the teaching signal for distillation,
and the student model with simpler architectures attempts to fit the task-specified fine-tuned BERT afterward.
Such  methodologies can achieve satisfying results by capturing the task-specified biases~\cite{mccallum1999text,godbole2018siamese,min2019compositional}, 
which are inherited by the tuned BERT~\cite{niven2019probing,DBLP:conf/acl/McCoyPL19}.
Unfortunately, the powerful generalization nature of BERT tends to be lost. 
Apparently, the original motivation of distilling BERT is to obtain a lightweight substitution of BERT for online implementations,
and the general semantic knowledge of BERT,
which plays a significant role in some NLP tasks like sentence similarity quantification, is expected to be maintained accordingly.
Meanwhile, for many NLP tasks, manual labeling is quite a high-cost work, and large amounts of annotated data can not be guaranteed to obtain. 
Thus, it is of great necessity to compress BERT with the non-task-specific training procedure on unlabeled datasets. 

For achieving the Non-task-specific Distillation from BERT, this paper proposes a distillation loss function to approximate sentence representations, by minimizing the cosine distance between the sentence representation given by the student network and the one from BERT. 
As a result, a student network with a much smaller scale of parameters is produced.
Since the distilling strategy purely focuses on the simulation of sentence embeddings from BERT, 
which is not directly related to any specific NLP task,
the whole training procedure takes only a large amount of sentences without any manual labeling work.
Same as BERT, the smaller student network can also perform transfer learning to any sentence-level downstream tasks.
The proposed methodology is evaluated on the open platform of General Language Understanding Evaluation (GLUE)~\cite{DBLP:conf/iclr/WangSMHLB19},
including the Single Sentence (SST-2), Similarity and Paraphrase (QQP and MRPC) and Natural Language Inference (MNLI) tasks.
The experimental results show that our proposed model outperforms the models distilled from a BERT fine-tuned on a specific task. 
Moreover, our model inferences more efficiently than other transformer-based distilled models.


\section{Related Works}
\label{sec:related}

With the propose of ELMo~\cite{peters2018deep}, 
various studies take the representation given by pre-trained language models as additional features to improve the performances.
~\newcite{howard2018universal} propose Universal Language Model Fine-tuning (ULMFiT), an effective transfer learning method that can be applied to any task in NLP, 
and accordingly, using pre-trained language models in downstream tasks became one of the most exciting directions.
On this basis, developing with deeper network design and more effective training methods, 
the performances of pre-trained models improved continuously~\cite{devlin2019bert,radford2019language,yang2019xlnet,liu2019roberta}.
Since the release of BERT~\cite{devlin2019bert}, 
the state-of-the-art (SOTA) results on 11 NLP tasks have been produced consequently.

With the improvement in performances, the computing cost increases, and the inference procedure becomes slower accordingly.
Thus, various studies focused on the model compression upon BERT.
Among the most common model compression techniques, the knowledge distillation~\cite{hinton2015distilling} has been proven to be efficient in transferring the knowledge from large-scaled pre-trained language models into another one~\cite{liu2019improving,barkan2019scalable,chia2019transformer}. 
With the help of proposed distillation loss, 
~\newcite{sun2019patient} compressed BERT into fewer layers by shortening the distance of internal representations between student and teacher BERTs.
For the sentence-pair modeling,
~\newcite{barkan2019scalable} found the cross-attention function across sentences is consuming and tried to remove it with distillation on sentence-pair tasks.
Different from these studies distilling BERT into transformer-based models,~\newcite{chia2019transformer} proposed convolutional student architecture to distill GPT for efficient text classification.
Moreover, focusing on the sequence labeling tasks,~\cite{tsai2019small} derived a BiLSTM or MiniBERT from BERT via standard distillation procedure to simulate the prediction on each token.
Besides,~\newcite{tang2019natural,tang2019distilling} proposed to distill BERT into a BiLSTM based model with penalizing the mean square error between the student's logits and the ones given by BERT as the objective on specific tasks, 
and introduced various data augmentation methods during distillation.


\begin{figure*}[htbp!]
    \centering
    \subfloat[Distilling BERT based on Representation Approximation.\label{fig:distill_arch}]{
        \includegraphics[width=.506\linewidth,angle=0]{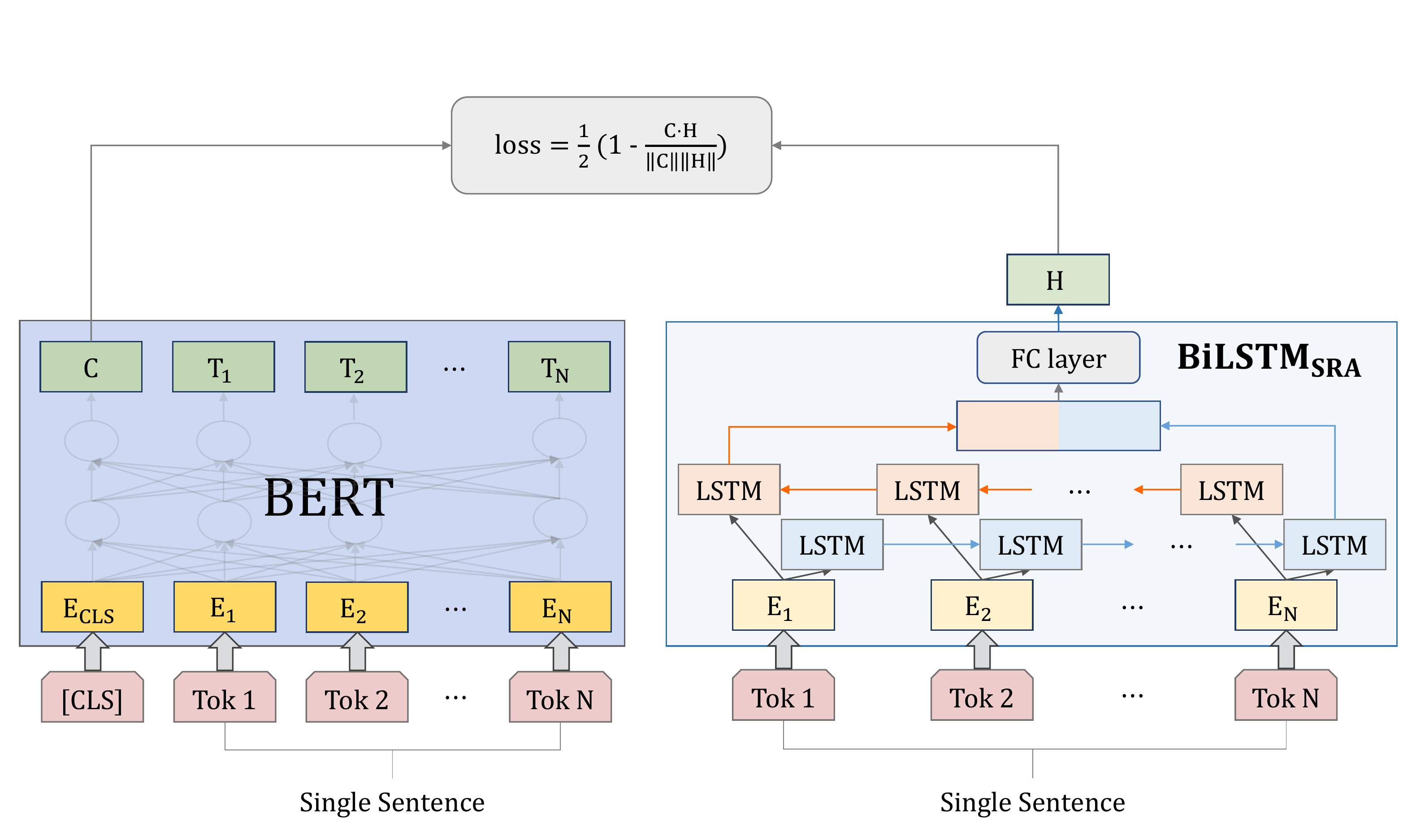}}\hfill
    \subfloat[Tuning (sentence classification).\label{fig:single_sen}]{
        \includegraphics[width=.158\linewidth,angle=0]{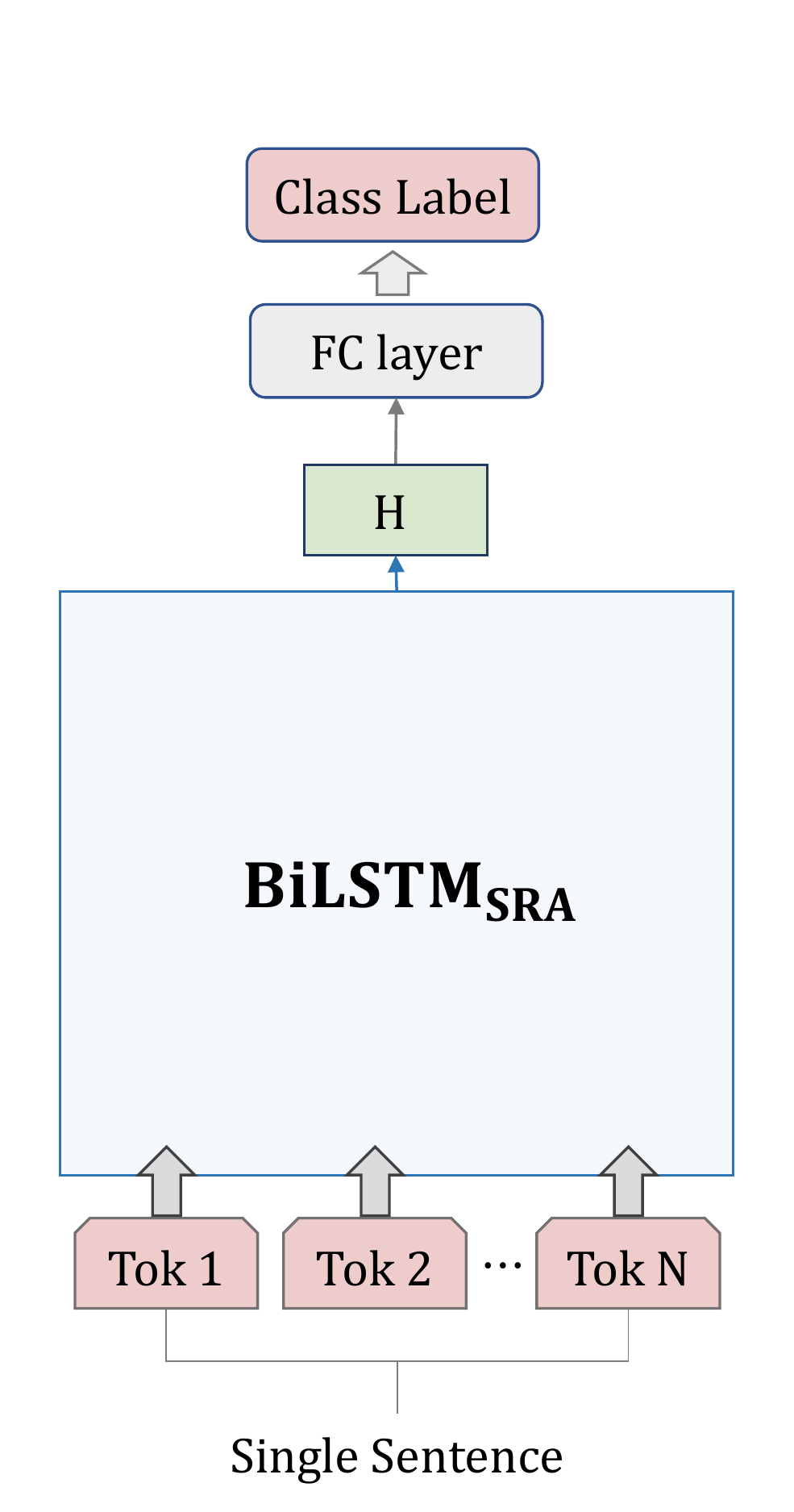}}\hfill
    \subfloat[Tuning (sentence-pair oriented).\label{fig:two_sen}]{
        \includegraphics[width=.316\linewidth,angle=0]{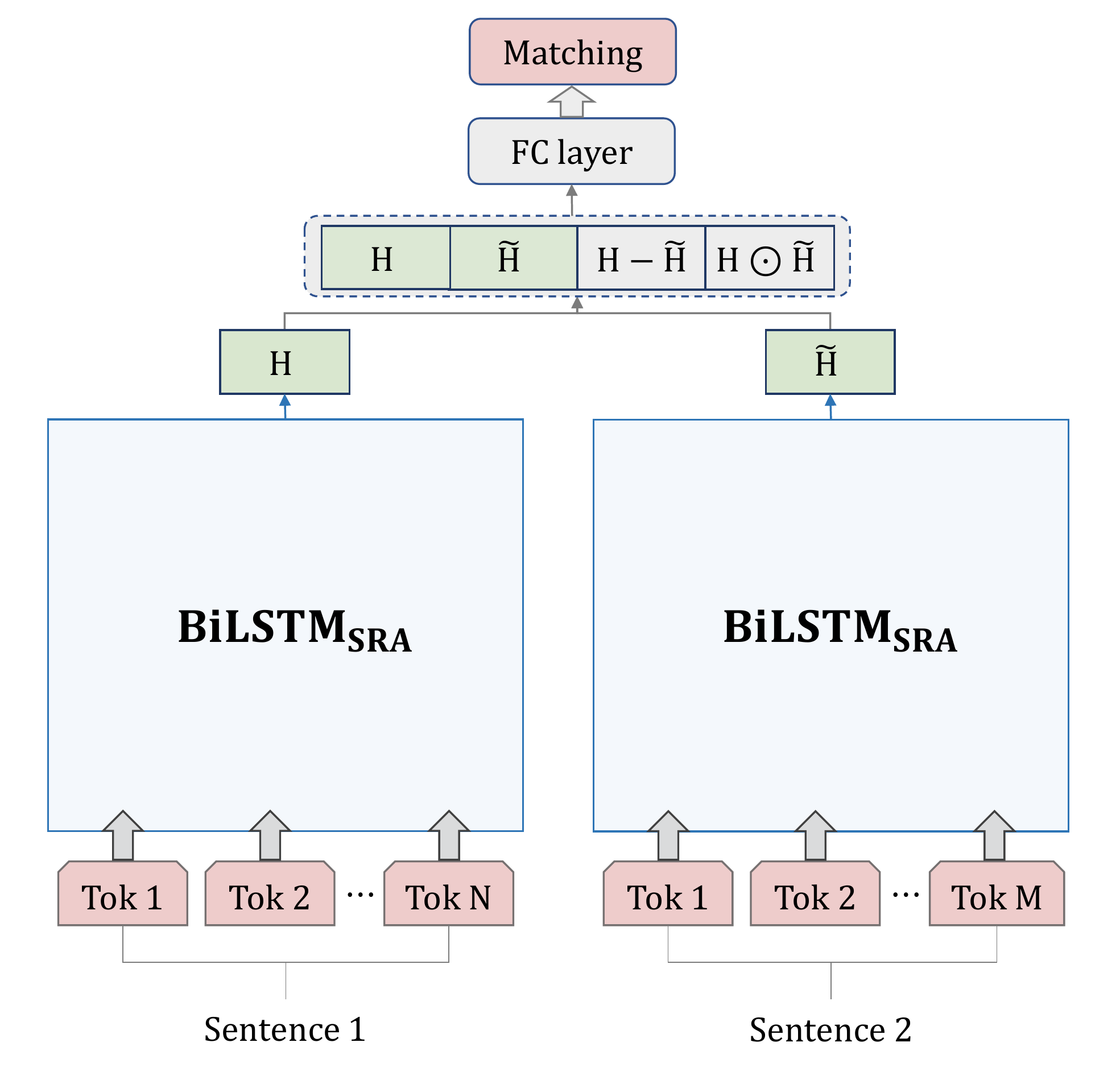}}
    \caption{The illustration of the proposed BERT distillation architecture including the distilling and tuning procedures. Sub-figure (a) demonstrates the distillation procedure taking BERT as the teacher model and BiLSTM as the student model, with the objective of approximating the representations given by BERT. (b) and (c) show two types of fine-tuning frameworks, in which (b) addresses the sentence classification task with the single sentence as the input, and (c) goes for the sentence-pair oriented tasks, i.e., sentence similarity quantification.}
    \label{fig:arch}
\end{figure*}

\section{Method}
\label{sec:method}

As introduced in Section~\ref{sec:intro}, our proposed method consists of two procedures. 
Firstly, we distill BERT into a smaller student model via approximating the representation of sentences given by BERT. 
Afterward, similar to BERT, the student model can be fine-tuned on any sentence-level task.

\subsection{Distillation Procedure}

Suppose $x=\{w_1,w_2,\cdots,w_i, \cdots w_n|i\in[1,n]\}$ stands for a sentence containing $n$ tokens (${w_i}$ is the $i$-th token of $x$),
and let $T:x \to T_x \in \mathbb{R}^d$ be the teacher model which encodes $x$ into $d$-dimensional sentence embedding $T_x$, 
the goal of the sentence approximation oriented distillation is to train a student model $S:x \to S_x \in \mathbb{R}^d$ generating $S_x$ as the approximation of $T_x$.

In our proposed distillation architecture, as shown in Figure~\ref{fig:distill_arch}, 
we take the BERT as the teacher model $T$, 
and the hidden representation $C$ is extracted from the top transformer layer upon the [CLS]\footnote{[CLS] is a special symbol added in front of other tokens in BERT, and the final hidden state corresponding to this token is usually used as the aggregate sequence representation.} token as $T_x$.
For the student model, a standard bidirectional LSTM (BiLSTM) is first employed to encode the sentence into a fixed-size vector $H$.
After that, a fully connected layer without bias terms is built upon the BiLSTM layer to map $H$ into a $d$-dimensional representation,
followed by a $tanh$ activation which normalizes the values in previous representation between -1 and 1 as the final $S_x$.

As our non-task-specific distillation task has no labeling data,
and the signal given by the teacher is a real value vector, it is not feasible to minimize the cross-entropy loss over the soft labels and ground truth labels~\cite{sun2019patient,barkan2019scalable,tang2019distilling}.
On this basis, we propose an adjusted cosine similarity between the two real value vectors $T_x$ and $S_x$ to perform the sentence representation approximation.
Our distillation objective is computed as follows:
\begin{equation}
\label{equ:loss}
\mathcal{L}_{distill} = \frac{1}{2} (1 - \frac{T_x \cdot S_x}{\|T_x\|\|S_x\|})
\end{equation}
Here $tanh$ is chosen as the activation function since most values (more than 98\% according to our statics) in $T_x$ obtained from BERT are within range of $tanh$ (-1 to 1).
The choice of using cosine similarity based loss is mainly based on the following two considerations.
Firstly, since 2\% values in $T_x$ are outside the range of [-1, 1], it is more reasonable to use a scalable measurement, such as cosine similarity, to deal with these deviations.
Secondly, it is meaningful to compute the cosine similarity between sentence embeddings given by BERT~\cite{xiao2018bertservice}.

Overall, after the distillation procedure, we obtained a BiLSTM based ``BERT'', which is smaller in parameter scale and more efficient in generating a semantic representation for a sentence.

\begin{description}[leftmargin=0cm]
\item[Distilling data] As our distillation procedure needs no dependency on sentence type or labeling resources but only standard sentences available everywhere, the distillation data selection follows the existing literature on language model pre-training as well as BERT.
We use the English Wikipedia to perform the distillation.
Furthermore, as the proposed method focus on the sentence representation approximation, the document is segmented into sentences using spacy~\cite{honnibal2017spacy}.
\end{description}

\subsection{Fine-tuning the Student Model}
The fine-tuning on sentence-level tasks is straightforward.
The downstream tasks discussed in this paper can be summarized as two types: type judgment on a single sentence and predicting the relationship between two sentences (same as all the tasks of GLUE). 
Figure~\ref{fig:single_sen} illustrates the model architecture for single sentence classification tasks.
The student model $S$ is utilized to provide sentence representation. 
After that, a multilayer perceptron (MLP) based classifier using Relu as activation of hidden layers is applied for the specific task.
For the sentence pair tasks, as shown in Figure~\ref{fig:two_sen}, the representations $H$ and $\tilde{H}$ for the sentence pair are obtained by transforming two sentences into two BiLSTM based student models with shared weights respectively.
Then, following the baseline BiLSTM model reported by GLUE~\cite{DBLP:conf/iclr/WangSMHLB19}, we apply a standard concatenate-compare operation between the two sentence embeddings and get an interactive vector as $[H, \tilde{H}, |H - \tilde{H}|, H \odot \tilde{H}]$, where the $\odot$ demotes for the element-wise multiplication.
Then, same as the single sentence task, an MLP based classifier is built upon the interactive representation.

For both types of tasks, MLP layers are initialized randomly, and the rest parameters are inherited from the distilled student model. 
Meanwhile, all parameters are optimized through the training procedure for the specific task.

\section{Experimental Setups}

\subsection{Datasets \& Evaluation Tasks}

To evaluate the performance of our proposed non-task-specific distilling method, we conduct experiments on three types of tasks: sentiment classification (SST-2), similarity (QQP, MRPC), and natural language inference (MNLI). 
All the tasks come from the GLUE benchmark~\cite{DBLP:conf/iclr/WangSMHLB19}.

\begin{description}[leftmargin=0cm]
\item[SST-2] Based on the Stanford Sentiment Treebank dataset~\cite{socher2013recursive}, 
the SST-2 task is to predict the binary sentiment of a given single sentence. 
The dataset contains 64k sentences for training and remains 1k for testing.

\item[QQP] The Quora Question Pairs\footnote{\url{https://www.quora.com/q/quoradata/First-Quora-Dataset-Release-Question-Pairs}} dataset consists of pairs of questions, 
and the corresponding task is to determine whether each pair is semantically equivalent.

\item[MNLI] The Multi-Genre Language Inference Corpus~\cite{williams2018broad} is a crowdsourced collection of sentence pairs with textual entailment annotations. 
There are two sections of the test dataset: matched (in-domain, noted as MNLI-m) and mismatched (cross-domain, noted as MNLI-mm).

\item[MRPC] The Microsoft Research Paraphrase Corpus~\cite{dolan2005automatically} is similar to the QQP dataset. This dataset consists of sentence pairs with binary labels denoting their semantic equivalence. 
\end{description}

\subsection{Model Variations}

\begin{description}[leftmargin=0cm]
\item[BERT~\cite{devlin2019bert}] with two variants: BERT$_{\text{BASE}}$ and BERT$_{\text{LARGE}}$, containing 12 and 24 layers of Transformer respectively.

\item[ELMO Baseline~\cite{DBLP:conf/iclr/WangSMHLB19}] is a BiLSTM based model, 
taking ELMo~\cite{peters2018deep} embeddings in place of word embeddings. 

\item[BERT-PKD~\cite{sun2019patient}] proposes a patient knowledge distillation approach to compress BERT into a BERT with fewer layers. 
BERT$_{3}$-PKD and BERT$_{6}$-PKD stand for the student models consisting of 3 and 6 layers of Transformer, respectively. 

\item[DSE~\cite{barkan2019scalable}] is a sentence embedding model based on knowledge distillation from cross-attentive models. 
For each single sentence modeling, the 24-layers BERT is employed.

\item[BiLSTM$_{\text{KD}}$~\cite{tang2019distilling}] introduces a new distillation objective to distill a BiLSTM based model from BERT for a specific task. BiLSTM$_{\text{KD}}$+TS~\cite{tang2019natural} donates the distilling procedure performed with the proposed data augmentation strategies.

\item[BiLSTM$_{\text{SRA}}$] stands for the Sentence Representation Approximation based distillation model proposed in this paper. BiLSTM$_{\text{SRA + KD}}$ donates performing knowledge distillation method proposed by~\newcite{tang2019distilling} during fine-tuning on a specific task, and BiLSTM$_{\text{SRA + KD}}$+TS demonstrates using the same augmented dataset to perform the distillation.
\end{description}

\begin{table*}[t]
  \centering
    \begin{tabular}{c l c c c c }
        \toprule
        \multirow{2}{*}{\#} & \multirow{2}{*}{Models} &  SST-2 & QQP & MNLI-m/mm & MRPC\\
        \cmidrule(lr){3-6}
        & & Acc & F$_{1}$/Acc & Acc & F$_{1}$/Acc\\\midrule
        1 & BiLSTM (report by GLUE)                           & 85.9 & 61.4\,/\,81.7 & 70.3\,/\,70.8 & 79.4\,/\,69.3\\
        2 & BiLSTM (report by~\newcite{tang2019distilling})   & 86.7 & 63.7\,/\,86.2 & 68.7\,/\,68.3 & 80.9\,/\,69.4 \\
        3 & BiLSTM (our implementation)                       & 84.5 & 60.3\,/\,81.6 & 70.8\,/\,69.4 & 80.2\,/\,69.7 \\
        4 & ELMO Baseline~\cite{DBLP:conf/iclr/WangSMHLB19}   & 90.2 & 65.6\,/\,85.7 & 72.9\,/\,73.4 & 84.9\,/\,78.0\\
        5 & BERT$_{\text{BASE}}$~\cite{devlin2019bert}        & 93.5 & 71.2\,/\,89.2 & 84.6\,/\,83.4 & 88.9\,/\,84.8 \\
        6 & BERT$_{\text{LARGE}}$~\cite{devlin2019bert}       & 94.9 & 72.1\,/\,89.3 & 86.7\,/\,85.9 & 89.3\,/\,85.4\\\hline
        7 & DSE~\cite{barkan2019scalable}                     & - & 68.5\,/\,86.9 & 80.9\,/\,80.4 & 86.7\,/\,80.7 \\
        8 & BERT$_{6}$-PKD~\cite{sun2019patient}              & 92.0 & 70.7\,/\,88.9 & 81.5\,/\,81.0 & 85.0\,/\,79.9\\
        9 & BERT$_{3}$-PKD~\cite{sun2019patient}              & 87.5 & 68.1\,/\,87.8 & 76.7\,/\,76.3 & 80.7\,/\,72.5\\\hline
        10 & BiLSTM$_{\text{KD}}$~\cite{tang2019natural}       & 88.4 & -\,/\,- & -\,/\,- & 78.0\,/\,69.7 \\
        11 & BiLSTM$_{\text{SRA}}$ \textbf{(Ours)}             & 90.0 & 64.4\,/\,86.2 & \textbf{72.6}\,/\,\textbf{72.5} & \textbf{83.1}\,/\,\textbf{75.1} \\
        12 & BiLSTM$_{\text{SRA + KD}}$      & \textbf{90.2} & \textbf{67.7}\,/\,\textbf{87.8} & 72.3\,/\,72.0 & 80.2\,/\,72.8 \\\hline
        13 & BiLSTM$_{\text{KD}}$+TS~\cite{tang2019distilling} & 90.7 & 68.2\,/\,88.1 & \textbf{73.0}\,/\,72.6 & 82.4\,/\,76.1 \\
        14 & BiLSTM$_{\text{SRA + KD}}$+TS   & \textbf{91.1} & \textbf{68.4}\,/\,\textbf{88.6} & \textbf{73.0}\,/\,\textbf{72.9} & \textbf{83.8}\,/\,\textbf{76.2} \\
        \hline \hline
        \multicolumn{6}{c}{Improvements obtained by performing different knowledge distillations} \\\hline
        15 & PKD~\cite{sun2019patient}    & +1.1 & +2.3\,/\,+0.9 & \textbf{+1.9}\,/\,+2.0 & +0.2\,/\,-0.1 \\
        16 & KD~\cite{tang2019natural}    & +1.7 & -\,/\,- & -\,/\,- & -2.9\,/\,+0.3 \\
        17 & SRA\textbf{(Ours)}        & +5.5 & +4.1\,/\,+4.6 & +1.8\,/\,\textbf{+3.1} & \textbf{+2.9}\,/\,\textbf{+5.4} \\
        18 & SRA\textbf{(Ours)}+KD     & \textbf{+5.7} & \textbf{+7.4}\,/\,\textbf{+6.2} & +1.5\,/\,+2.6 & 0.\,/\,+3.1 \\\hline
        19 & KD+TS~\cite{tang2019natural}    & +4.0 & +4.5\,/\,+1.9 & \textbf{+4.3}\,/\,\textbf{+4.2} & +1.5\,/\,\textbf{+6.7} \\
        20 & SRA\textbf{(Ours)}+KD+TS  & \textbf{+6.6} & \textbf{+8.1}\,/\,\textbf{+7.0} & +2.2\,/\,+3.5 & \textbf{+3.6}\,/\,+6.5 \\
        \bottomrule
  \end{tabular}%
  \caption{Evaluation results with scores given by the official evaluation server\protect\footnotemark.}
  \label{tbl:overall}
\end{table*}

\subsection{Hyperparameters}

For the student model in our proposed distilling method, 
we employ the 300-dimension GloVe (840B Common Crawl version;~\citealp{pennington2014glove}) to initialize the word embeddings.
The number of hidden units for the bi-directional LSTM is set to 512,
and the size of the task-specific layers is set to 256.
All the models are optimized using Adam~\cite{DBLP:journals/corr/KingmaB14}. 
In the distilling procedure, we choose the learning rate as $1\times10^{-3}$ with the batch size=1024. 
During fine-tuning, the best learning rate on the validation set is picked from $\{2,3,5,10\}\times10^{-4}$.
For the data augmentation, we use the rule-based method originally suggested by~\newcite{tang2019distilling}.
Notably, on the SST-2 and MRPC dataset, we stop data augmenting when the transfer set achieves 800K samples following the setting of their follow-up research~\cite{tang2019natural}.
Besides, inspired by the comparisons in the research of~\newcite{sun2019patient}, we find BERT$_{\text{BASE}}$ can provide more instructive representations than BERT$_{\text{LARGE}}$.
So that, we chose BERT$_{\text{BASE}}$ as our teacher model to pre-train the BiLSTM$_{\text{SRA}}$.

\section{Results and Analysis}

\subsection{Model Performance Analysis}

\footnotetext{\url{https://gluebenchmark.com/leaderboard}}

For a comprehensive experiment analysis, we collect data and implement comparative experiments on various published BERT and BERT-distillation methods.
Table~\ref{tbl:overall} shows the results of our proposed BiLSTM$_{\text{SRA}}$ and the baselines on the four datasets.
All models in the first block (row 1-6) belong to base methods without implementing distillation, 
the second (row 7-9) and third (row 10-12) blocks show the performances of distillation models using BERT and BiLSTM structures respectively,
while the forth block (row 13-14) displays the influences of textual data augmentation approach on our BiLSTM$_{\text{SRA}}$ and BiLSTM$_{\text{KD}}$ distillation baseline.
The last two blocks contain the results of pure improvements obtained by different distillation methods.
To analyse the effectiveness of BiLSTM$_{\text{SRA}}$ thoroughly, we break down the analyses into the following two perspectives.

\subsubsection{Comparison Between Models}
Taking those non-distillation methods in the first block as references, 
BiLSTM$_{\text{SRA}}$ performs on par with ELMO on all tasks.
Especially, BiLSTM$_{\text{SRA + KD}}$+TS outperforms the ELMO baseline by approximately 3\% on QQP and 1\% on SST-2 (row 14 vs 4). 
Such fact shows our compressed ``BERT'' can provide as good pre-trained representations as those given by ELMO on the sentence-level tasks.

For those distillation methods, both our model and BiLSTM$_{\text{KD}}$ distill knowledge from BERT into a simple BiLSTM based model, 
while BERT-PKD focus on distilling with the BERT of fewer layers.
Despite of the powerful BERT based student model and large-scale parameters used by BERT-PKD, our proposed BiLSTM$_{\text{SRA}}$ still outperforms BERT$_{3}$-PKD on SST-2 and MRPC dataset  (row 12 vs 9).
For BiLSTM$_{\text{KD}}$, it proposes a rule-based textual data augmentation approach (noted as TS) to construct transfer sets for the task-specific knowledge distillation.
We also employ such method upon BiLSTM$_{\text{SRA + KD}}$.
With and without the data augmentation, BiLSTM$_{\text{SRA}}$ consistently outperforms BiLSTM$_{\text{KD}}$ on all tasks (row 12 vs 10; row 14 vs 13).
Collaborating with the standard knowledge distillation and data augmentation methods, our proposed model is sufficient to distill semantic representation modeled from pre-training tasks as well as the task-specific knowledge included in a fine-tuned BERT.

Besides, for the sentence matching task, the overall architecture of DSE is similar to our model except DSE employs the pre-trained BERT$_{\text{LARGE}}$ to give sentence representations.
Thus, on the sentence-pair level tasks, DSE somehow is an upper bound of the distilled models without utilizing any cross attention to model the interaction between the two sentences.
Comparing with DSE achieved an averaged 80.7 score on all sentence-pair level tasks, BiLSTM$_{\text{SRA + KD}}$+TS can also obtain 77.2 that only 3.5 points lower (row 7 vs 14). 
Analyzing from this fact, our proposed model has distilled a much smaller ``BERT'' with acceptable performances.

\subsubsection{Distillation Effectiveness}
Because in each paper, the performances of student models used for distillation vary from each other.   
To further evaluate the distillation effectiveness, we also report the improvement of each distillation method upon the corresponding student directly trained without distillation (in row 15-20).
It can be observed that SRA improves the scores by over 3.9\% on average, while PKD and KD only provide less than 1.2\% increase (row 17/16 vs 15).

\begin{table}[ht]
  \centering
    \begin{tabular}{l c c }
        \toprule
        Models &  \# of Par. & Inference Time \\
        \midrule
        BERT$_{\text{LARGE}}$   & 309\,(64x) & 1461.9\,(54.4x)  \\
        BERT$_{\text{BASE}}$    & 87\,(18x) & 479.7\,(17.7x)  \\ 
        ELMO                    & 93\,(19x)  & -\,(23.7x)  \\\hline
        BERT$_{3}$-PKD          & 21\,(4x)  & -\,(4.8x)  \\
        BERT$_{6}$-PKD          & 42\,(9x)  & -\,(9.2x)  \\ 
        DSE                     & 309\,(64x) & -\,(109.1x)  \\ 
        BiLSTM$_{\text{KD}}$    & \textbf{2.4\,(0.5x)}  & 31.9\,(1.2x) \\
        BiLSTM$_{\text{SRA}}$   & 4.8\,(1x)  & \textbf{26.8\,(1x)}  \\
        \bottomrule
  \end{tabular}%
  \caption{Comparisons of model size and inference speed. \# of Par. denotes number of millions of parameters and the inference time is in seconds. The factors inside the brackets are computed comparing to our proposed model.}
  \label{tbl:efficiency}
\end{table}

Since our distillation method is unrelated to specific tasks, 
KD can also be performed upon BiLSTM$_{\text{SRA}}$ during fine-tuning on a given dataset.
This operation provides a notable boost on QQP task, but damages the performance on both MNLI and MRPC datasets (row 17 vs 18).
We attribute these differences to the following aspects:
a) the QQP dataset has more obvious task-specified biases during the sampling process\footnote{\url{https://www.kaggle.com/c/quora-question-pairs/discussion/32819\#latest-189493}}.
A pre-trained BERT can not learn such biases; 
b) a fine-tuned BERT on the MNLI can not further provide more easy-to-use information to guide the student training after performing SRA;
c) MRPC does not include enough data to complete KD,
which is also indicated by the decreased F1 score shown in row 16 in Table~\ref{tbl:overall}.
These phenomena reflect that the pre-distillation without paying attention to a specific task can help to learn more useful semantic information from the teacher model.

Different from obtaining the best results on the MNLI dataset, SRA+KD+TS brings few improvements comparing to KS+TS (row 19 vs 20).
We attribute this to the difference in the results of pure student BiLSTM between our implementation and the one of~\newcite{tang2019distilling}, though our scores are more constant with the baselines given by the GLUE benchmark~\cite{DBLP:conf/iclr/WangSMHLB19}.

\subsection{Model Efficiency Analysis}

To compare the inference speeds of different models, 
we also implement experiments on 100k samples from the QQP dataset,
the results are shown in Table~\ref{tbl:efficiency}. 
The inference procedure is performed on a single P40 GPU with a batch size of 1024.
As the inference time is affected by the computing power of the test machine, 
for fair comparisons with ELMO, BERT$_{3}$-PKD, BERT$_{6}$-PKD and DSE, 
we inherit the speed-up factors from previous papers.
Besides, the numbers of parameters reported in Table~\ref{tbl:efficiency} exclude those from the embedding layers,
since such components do not affect the inference speed and are highly related to the vocabulary sizes, i.e., usually few words appeared for a specific task.


From the results shown in Table~\ref{tbl:efficiency},
it can be observed that the BiLSTM based distilled models have fewer parameters than BERT, ELMO, as well as the other transformer-based models.
Both the BERT$_{\text{BASE}}$ and ELMO are around 20 times larger in parameter size and 20 times slower in inference speed.
Even the smallest transformer based model BERT$_{3}$-PKD is also four times larger than our proposed BiLSTM$_{\text{SRA}}$.
Comparing with BiLSTM$_{\text{KD}}$, although our proposed BiLSTM$_{\text{SRA}}$ is larger in parameter size due to the restriction of the sentence embedding's dimension given by the teacher BERT, it stills inferences more efficiently.
This is mainly due to the fact that the more hidden units in BiLSTM$_{\text{SRA}}$ are more accessible to calculated in parallel by the GPU core,
while the larger word embedding size in BiLSTM$_{\text{KD}}$ slows down its inference efficiency.
In conclusion, the cost and production per second of BiLSTM$_{\text{KD}}$ and BiLSTM$_{\text{SRA}}$ are within the same scale, 
but our method achieves better results on GLUE tasks according to the comparison shown in Table~\ref{tbl:overall}.

\subsection{Influence of Task-specific Data Size}
\label{sec:few_data}

Since pre-trained language models have well-initialized parameters and only learn a few parameters from scratch, these models usually converge faster and are less dependent on large-scale annotations.
Correspondingly, the non-task-specific distillation method proposed in this paper also aims at obtaining a compressed pre-trained BERT as well as keeping these desirable properties.
To evaluate it, in this section, we discuss the influence of the task-specific training data and learning iterations on the performance of our model and the others.

\begin{figure}
    \centering
    \includegraphics[height=1.\linewidth,angle=270]{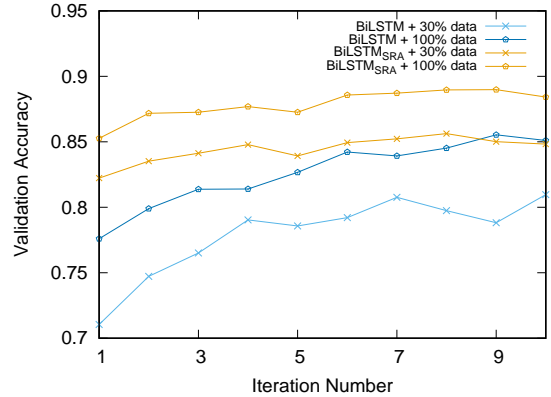}
    \caption{Learning curve on the QQP dataset.}
    \label{fig:learning_curve_qqp}
\end{figure}
\begin{table}[!t]
  \centering
    \begin{tabular}{l c c c c}
        \toprule
        Models &  20\% & 30\% & 50\% & 100\% \\
        \midrule
        BERT$_{\text{LARGE}}$    & 91.9 & 92.5 & 93.5 & 93.7 \\\hline
        BiLSTM                   & 80.7 & 81.0 & 83.6 & 84.5 \\
        BiLSTM$_{\text{KD}}$     & 81.9 & 83.2 & 84.8 & 86.3 \\
        BiLSTM$_{\text{SRA}}$    & \textbf{85.9} & \textbf{87.3} & \textbf{88.1} & \textbf{89.2} \\
        \bottomrule
  \end{tabular}%
  \caption{The accuracy scores evaluated on the SST-2 validation set. The models are trained with different proportions of the training data.}
  \label{tbl:few}
\end{table}

As illustrated in Table~\ref{tbl:few}, we experiment in training the models using different proportions of the dataset.
BERT$_{\text{LARGE}}$ trained on the corresponding data stands for the teacher model of each BiLSTM$_{\text{KD}}$.
With no doubt, all the models can achieve better results using more training data, while BERT performs the best. 
BERT even successfully predicts 91.9\% of validation samples under only 20\% training data.
Comparing with the pure BiLSTM models, the BiLSTM$_{\text{KD}}$ models slightly improve the performances by 1\%$\sim$2\%,
whereas BiLSTM$_{\text{SRA}}$ outperforms the best BiLSTM model as well as the BiLSTM$_{\text{KD}}$ trained with 20\% and 30\% percent data respectively.
Besides, similar to BERT, the difference of accuracy between BiLSTM$_{\text{SRA}}$ trained with 20\% and the one using 100\% corpus is relatively small.
This phenomenon indicates that our model converges faster and is less dependent on the amount of training data for downstream tasks.


Such conclusions are also reflected in the comparison in Figure~\ref{fig:learning_curve_qqp} of the models' learning curves on QQP.
Even though QQP is a large dataset to train a good BiLSTM model, it can be observed that BiLSTM$_{\text{SRA}}$ trained with 30\% data performs equivalent to BiLSTM using the whole corpus.
Moreover, using 100\% training data, BiLSTM$_{\text{SRA}}$ even outperforms the converged BiLSTM after the first epoch.
Besides, all the BiLSTM$_{\text{SRA}}$ models converge in much fewer epochs.

\begin{table}[ht]
  \centering
    \begin{tabular}{c c c c}
        \toprule
        \multirow{2}{*}{Size} & Distillation & SST-2 & MNLI-m\\
        \cmidrule(lr){2-4} 
        & Loss & Acc & Acc\\\midrule
        0M  & -      & 84.5        & 70.23 \\
        1M  & 0.0288 & 88.9\,(+4.4)  & 72.01\,(+1.78) \\  
        2M  & 0.0257 & 89.3\,(+4.8)  & 72.09\,(+1.86) \\
        4M  & \textbf{\textit{0.0241}} & \textbf{89.4\,(+4.9)} & \textbf{72.45\,(+2.22)} \\
        \bottomrule
  \end{tabular}%
  \caption{The distillation losses on the Wikipedia validation set and the accuracy scores of the downstream tasks various with the distillation data sizes.}
  \label{tbl:few_distill}
\end{table}

\subsection{Influence of Distilling Data Size}
\label{sec:distill_data_size}

Despite the task-specified data, 
Wikipedia corpus is used in the distillation procedure of our proposed method.
We also train different BiLSTM$_{\text{SRA}}$ base models using \{1, 2, 4\} million Wikipedia data, 
and the corresponding fine-tuning performances on SST-2 and MNLI are reported in Table~\ref{tbl:few_distill}.
It can be observed that both the performances of BiLSTM$_{\text{SRA}}$ on SST-2 and MNLI are proportional to the distillation loss.
This observation indicates the effectiveness of our proposed distillation process and objective. 
Besides, distilling with adequate data is sufficient to produce more BERT-like sentence representations.
Nevertheless, different from the fact that more training data has a significant benefit in a particular task, four times the distilling data only improve around 0.5 points on both two downstream tasks.
Thus, our method does not require a vast amount of training data and a long training time to obtain good sentence representations.
Furthermore, the loss scores in the second column suggest BiLSTM$_{\text{SRA}}$ can generate more than 95\% similar sentence embeddings with the ones given by BERT under the measure of the cosine distance.

\subsection{Analysis on the Untuned Sentence Representations}

A notable characteristic of the pre-trained language models, such as ELMO, BERT, and certainly the non-task oriented distillation models, 
lies in the capability of providing sentence representations for quantifying similarities of sentences, without any tuning operation based on specific tasks. 
In this subsection, we conduct the comparisons among models by directly extracting their sentence embeddings without fine-tuning upon sentence similarity oriented tasks. 

\begin{table}[!ht]
  \centering
    \begin{tabular}{l c c }
        \toprule
        Models &  Acc & F$_{1}$ \\
        \midrule
        ELMO                             & 65.1 & \textbf{64.4}  \\
        BERT$_{\text{BASE}}$ (CLS)       & 63.9 & 61.0           \\
        BERT$_{\text{BASE}}$ (averaged)  & \textbf{66.4} & 64.1  \\\hline
        BiLSTM$_{\text{KD}}$       & 56.3 & 56.6           \\
        BiLSTM$_{\text{SRA}}$      & \textbf{62.9} & \textbf{61.0}  \\
        \bottomrule
  \end{tabular}%
  \caption{Results of untuned sentence representing models on QQP dataset.}
  \label{tbl:unsupervised}
\end{table}

Table~\ref{tbl:unsupervised} lists the results of models on the QQP dataset. 
It should be noted that, in this table, 
ELMO, BERT$_{\text{BASE}}$ (CLS) and BERT$_{\text{BASE}}$ (averaged) are introduced as the comparison basis, 
since they can give the SOTA untuned sentence representations for the similarity measurement. 
The comparison mainly focuses on the performances of our proposed BiLSTM$_{\text{SRA}}$ and BiLSTM$_{\text{KD}}$. 
For a thorough comparison, 
we define the training objective of BiLSTM$_{\text{KD}}$ as fitting the cosine similarity score of the sentence pair directly given by the pre-trained BERT,
which means both the teacher BERT and KD do not utilize the labels of QQP dataset.
It can be seen that, even though the training goal of BiLSTM$_{\text{KD}}$ is more direct than BiLSTM$_{\text{SRA}}$,
our BiLSTM$_{\text{SRA}}$ outperforms the former on both the metrics,
and meanwhile achieves scores closed to those of BERT$_{\text{BASE}}$.
Besides, we can also observe that, for sentence similarity quantification, 
averaging the context word embeddings as the sentence representation (ELMO and  BERT$_{\text{BASE}}$ (averaged)) works better than taking the final hidden state corresponding to the [CLS] token (BERT$_{\text{BASE}}$ (CLS)).

\section{Conclusions}


In this paper, we have presented a sentence representation approximating oriented method for distilling the pre-trained BERT model into a much smaller BiLSTM without specifying tasks, 
so as to inherit the general semantic knowledge of BERT for better generalization and universal-usability.
The experiments conducted based on the GLUE benchmark have shown that our proposed non-task-specific distillation methodology can improve the performances on multiple sentence-level downstream tasks.
From the experimental results, the following conclusions can be drawn:
1) our proposed distillation method can bring the 5\% improvement to the pure BiLSTM model on average;
2) the proposed model can outperform the state-of-the-art BiLSTM based pre-trained language model which contains much more parameters;
3) compared to the task-specific distillation, our distilled model
is less dependent on the corpus size of the downstream task with satisfying performances guaranteed.


\bibliography{acl2019}
\bibliographystyle{acl_natbib}

\end{document}